# Multimodal Machine Translation with Reinforcement Learning

**Xin Qian**
School of Computer Science
Carnegie Mellon University
xinq@cs.cmu.edu

**Ziyi Zhong**
School of Computer Science
Carnegie Mellon University
ziyiz2@cs.cmu.edu

**Jieli Zhou**
School of Computer Science
Carnegie Mellon University
jieliz@cs.cmu.edu

## Abstract

Multimodal machine translation is one of the applications that integrates computer vision and language processing. It is a unique task givent that in the field of machine translation, many state-of-the-arts algorithms still only employ textual information. In this work, we explore the effectiveness of reinforcement learning in multimodal machine translation. We present a novel algorithm based on the Advantage Actor-Critic (A2C) algorithm that specifically cater to the multimodal machine translation task of the EMNLP 2018 Third Conference on Machine Translation (WMT18). We experiment our proposed algorithm on the Multi30K multilingual English-German image description dataset and the Flickr30K image entity dataset. Our model takes two channels of inputs, image and text, uses translation evaluation metrics as training rewards, and achieves better results than supervised learning MLE baseline models. Furthermore, we discuss the prospects and limitations of using reinforcement learning for machine translation. Our experiment results suggest a promising reinforcement learning solution to the general task of multimodal sequence to sequence learning.

## 1 Introduction

Machine Translation research has long been focusing on utilizing textual information alone, in which neural translation models have achieved good results for many languages, for example Jean et al. [8] presented a neural translation model between English and German and achieved state-of-the-art performance. In a recent trend, significant research has been done to combine signals from language and vision for joint modeling. Multimodal machine translation is the task that translates a source sentence into target language with the additional input from non-textual modal, e.g. images, videos, audios, etc. Specifically in this work, our translation task is built from an image captioning dataset, in which the image channel alone could be viewed as an image captioning task. In the field of image captioning, a common architecture is to use a convolutional neural network image feature encoder followed by a recurrent neural network textual sequence decoder, e.g. neural image caption (NIC) model by Vinyals et.al [12]. Xu et al. [13] further incremented this architecture with visual attention mechanism. The residual network (ResNet) proposed by He et al. [6] is a good candidate for such pre-trained image feature encoder.

In this project, we explore the area of multimodal machine translation, where not only source and target text are involved but also the corresponding. In our experiment, we focus on an image enhanced machine translation task, where the text to be translated is the caption of an image, as shown in Figure 1. And in this case, the image itself could be helpful in supervising the translation task. In later sections, we will present and Advantage Actor-Critic (A2C) approach in this multimodal translation task. And we will give a thorough evaluation and discussion on the result of this approach.

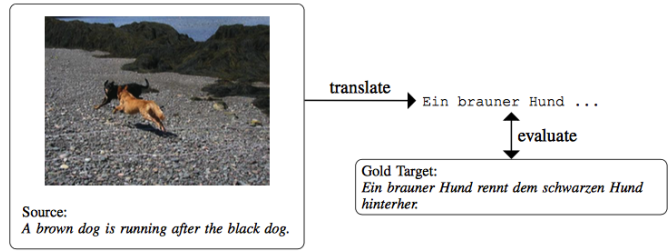

**Figure 1:** WMT18 multimodal machine translation task

Although applying reinforcement learning in machine translation task might seem counter-intuitive at first sight, there are two major motivations. First, the model shall be optimized towards the general target of improving system effectiveness, i.e., translation quality, as reflected by automatic evaluation metrics or human judgment. In the context of machine translation, for example, BLEU score is an effective metric for translation brevity and preciseness. The traditional encoder-decoder machine translation models, however, maximizes the log-likelihood of the current output given the previous output, which loses sight of the overall evaluation targets. To eliminate the training target discrepancy, we assign end evaluation metrics (which are typically not log-likelihood) as the reinforcement learning training reward,

Second, traditional machine translation models was known to have an *exposure bias* [2] between training and testing phases. At testing phase, the trained model conditions its future outputs on its own past guesses. When its own guess goes wrong, any future predictions are prone to the compounding errors and becomes problematic for outputting longer sequences.

To summarize, the main contributions of our paper are:

- a novel multimodal machine translation architecture that employs A2C model from reinforcement learning;

- a through analysis over the strength and weakness of the proposed model.

## 2    Related Work

In recent years, reinforcement learning has gained much popularity in sequence prediction tasks. Ranzato et al. [10] presented a algorithm using REINFORCE for machine translation and achieved state-of-the-art performance. Bahdanau et al. [1] presented an actor-critic algorithm for general sequence prediction and specifically machine translation, and use BLEU score as one choice of reward in this reinforcement learning process. However, these papers involve text-only machine translation.

The Conference of Machine Translation (WMT'16[1], WMT'17, WMT18) has hold Multimodal Translation Tasks for a consecutive 3 years (from 2016 to 2018). The task has been consistently aiming at generating image descriptions in target languages. Specifically, the multimodal machine translation task aims at translate English sentences that describe an image into German or French or Czech, given the English sentence itself and the image that it describes. In other words, the task can be decomposed into a translation task and a image captioning task. In WMT16 multimodal machine translation shared task [11], Calixto et al. from DCU proposed an attention-based neural machine translation model [3]. Guasch et al. presented a bidirectional RNN model with double embedding [5]. However, to the best of our knowledge, none of these models use reinforcement learning.

---

## 3 Methods

### 3.1 Problem Formulation

We formulate the problem of multimodal machine translation as a reinforcement learning problem with advantage actor-critic (A2C) methods as follows. This training procedure addresses the *exposure bias* problem from log-likelihood (MLE) training and ensures the model is trained on the metrics that we cared about. The general machine translation task generates an output target sentence $Y = (y_1, ..., y_T), y_t \in A$ (where $A$ is the target dictionary) given an input sequence of $X = (x_1, ..., x_T)$. At each timestamp, the *actor*, in our case, the encoder-decoder, applies a stochastic *policy* to get the next token.

- **State**: Source sentence $X = (x_1, ..., x_T)$ and generated subsequences $Y = (y_1, ..., y_{t-1})$.

- **Action**: Choose the next target token $y_t$ from the action space $A$, target dictionary.

- **Policy**: $p(y_t|x_1, ..., x_T, y_1, ..., y_{t-1})$.

- **Reward**: task evaluation metrics score of the generated subsequence, up to timestamp $t$, $r_t(\hat{Y}_{1..t}, Y)$.

### 3.2 Proposed Method

Our multimodal machine translation model uses an encoder-decoder architecture with global attention, as shown in figure 1. The encoder has two channels of input, the text and the image. The text encoder channel (also the decoder) is a recurrent neural network (RNN). Input source text sequences are represented by one-hot vectors and fed into the text encoder channel. The image encoder channel inputs raw image into pre-trained ResNet and extracted image features of 2048 dimensions. We concatenate image encoder outputs and text encoder outputs, feed them into an MLP layer, and then feed the output of the MLP layer into the decoder. The decoder outputs a sequence of state vectors, on top of which is the stochastic output layer (softmax) that generates discrete output target words.

In A2C training procedure, the model is the actor (Figure 2). The critic architecture greatly resembles the actor architecture. As shown in Figure 3, the only difference is critic output: at each time-stamp, it outputs a state value vector of size 1. Critic is trained with actor-generated episodes, of samples (predicted target words at each time-stamp) and rewards (e.g. a list of BLEU scores at each time-stamp, evaluated respectively on sub-sentences of $y_{1..1}, y_{1..2}, ..., y_{1..t}$). The step-wise rewards are treated as the ground-truth and MSE loss is used to train critic state value prediction. This implementation trick is known as *reward shaping* [1].

During testing time (aka, translation/decoding phase), we produce the target sentences given the source sentence with greedy search. Although an approximate beam search is usually conducted to approximate global optimal prediction, we implemented greedy search for decoding since the performance gap between the two methods is limiting in our case.

## 4 Experiments

### 4.1 Dataset Description

Multi30K [4] is a multilingual English-German image description dataset that stems from Flickr30K, which is the textual dataset for this task. It contains 31,014 German translations of English descriptions and 155,070 independently collected German descriptions. Translations are created by professionals, whereas image descriptions are crowd-sourced by non-professional workers. Along with this dataset, there is also a data preprocessing script that simplify the process of cleaning and preprocessing data. Additionally, there is Flickr30K Entity dataset, which is an extension of Flickr30K dataset that contains additional layers of annotations. WMT 18 also provides image features, which is extracted by a 50 layer Deep Residue Network (ResNet-50) [7]. This image feature data provides us useful image features that could directly be used for reinforcement learning.

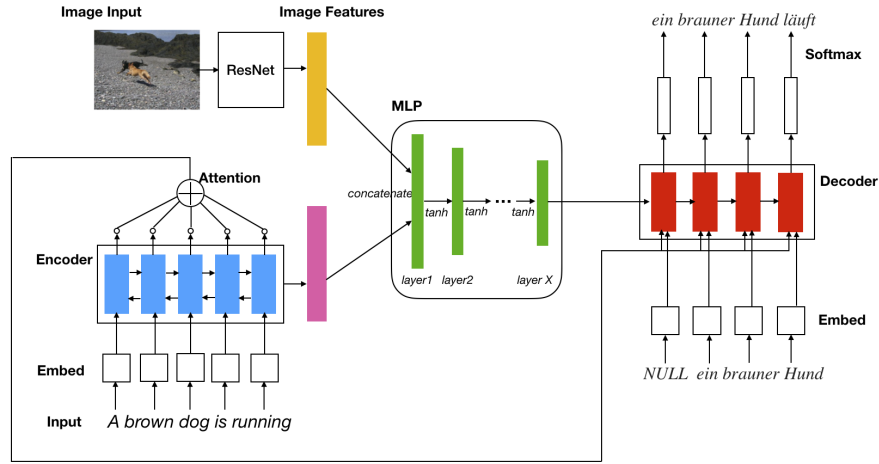

**Figure 2:** Actor for multimodal machine translation

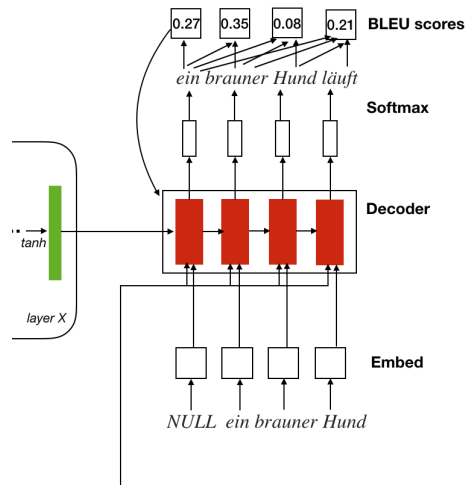

**Figure 3:** Critic for multimodal machine translation

For official evaluation, there will be an in-domain test on 1,071 test instances, which will be released on May 14th, 2018. Therefore we experiment our model on WMT17 multimodal machine translation shared task data. In this dataset, for each language pair (English-Czech, English-French, English-German), there are 29,000 training examples, 1000 validation examples and 1071 test examples. The vocabulary size for English, Czech, French and German are 10214, 22400, 11223, 18726 respectively.

## 4.2 Experiment Setup

We implemented the attention-based encoder-decoder architecture of actor and critic with supervised MLE training and A2C training. For encoder-decoder, we use a bidirectional LSTM with MLP-based global attention, both 256 dimensions of embedding unit and hidden unit. Training is done in batches of size 64, with Adam optimizer and decaying learning rate that starts from 1e-3. The target translation is generated with greedy search. All experiments were run on a p2.xlarge AWS instance.

### 4.3 Evaluation Metrics

**BLUE** We use BLEU score [9] (ngram of up to 4) as our major automatic evaluation metrics. BLEU score is the n-gram overlap between output translation and reference translation. And a brevity term is added to penalize short translations.

$$\text{BLEU} = \min(1, \frac{\text{output-length}}{\text{reference-length}})(\prod_{i=1}^{n} \text{precision}_n)^{\frac{1}{n}}$$

BLEU score has been widely applied to machine translation and image captioning tasks. When we trained our A2C model, we only used the sentence-level BLEU score to measure rewards for each timestamp, while at testing time, we also report the final model performance along with model perplexity and corpus-level BLEU score. Corpus-level BLEU scores accounts for the micro-average precision and is a unified measurement for corpus-level translation quality.

**Perplexity:** Perplexity is a measurement of how well our translation model predicts a sample. It is expressed as

$$\text{Perplexity} = \exp(\sum_x p(x) \log_e \frac{1}{p(x)})$$

**Other:** Outside the scope of this report, but potentially in real world crowdsourced machine translation setting, it is reasonable to replace BLEU with any human-judged translation quality metrics, e.g. in grade level 1-5, including: (1) adequacy, whether the target sentence has consistent semantic meaning as the source sentence; (2) fluency, whether the target sentence is in good fluent English. Users may also replace BLEU with customized automatic evaluation metrics, a syntactic soundness score from off-the-shelf dependency parser. [2].

### 4.4 Results and Analysis

#### 4.4.1 Overall performance

**Table 1:** Result for text-only translation

|  | Perplexity | | | Sentence-level Reward | | | **Corpus-level reward** | | |
|---|---|---|---|---|---|---|---|---|---|
|  | de | fr | cs | de | fr | cs | de | fr | cs |
| MLE | **8.83** | **3.83** | **16.05** | 41.67 | 57.27 | 38.38 | 35.23 | 52.49 | 27.58 |
| A2C wo pre-trained critic | 11.16 | 4.45 | 20.65 | **43.21** | **59.07** | **38.94** | **36.92** | **54.73** | **28.23** |
| A2C w/ pre-trained critic | 10.86 | 4.50 | 16.05 | 42.86 | 58.62 | 38.38 | 36.52 | 54.41 | 27.58 |

**Table 2:** Result for translation with image channel

|  | Perplexity | | | Sentence-level Reward | | | **Corpus-level reward** | | |
|---|---|---|---|---|---|---|---|---|---|
|  | de | fr | cs | de | fr | cs | de | fr | cs |
| MLE | **8.69** | **3.90** | **16.03** | 41.31 | 57.59 | 38.31 | 34.60 | 53.08 | 27.40 |
| A2C wo pre-trained critic | 11.38 | 4.67 | 21.04 | 42.30 | 58.19 | **38.78** | 35.67 | 53.81 | **28.21** |
| A2C w/ pre-trained critic | 11.36 | 4.57 | 22.30 | **43.36** | **58.19** | 38.68 | **37.12** | **54.02** | 28.05 |

Table 1 and 2 show the performance of different models after respective number of epochs of training. "MLE" is the supervised training setting that optimizes with log-likelihood for 40 epochs, which involves no reinforcement learning. "A2C wo pre-trained critic" was trained from 20 epochs of MLE, followed by 20 epochs of joint training. "A2C w/ pre-trained critic" was trained from 20 epochs of MLE, 20 epochs of pre-training critic, and 20 epochs of joint training. We consider this as a fair comparison to green curve and blue curve, since we fixed-actor (actual performance) while pre-training critic during epoch 21-40.

Corpus-level reward, which is the corpus-level BLEU score, is what we care most about. As for another metric, perplexity in its formula form, is very similar to MLE training loss function. It is

---

unsurprising that the model trained with MLE has the lowest preplexity score for all language pairs. As we can see in the table, for all three language pairs, using A2C (pre-trained critic or not) increases the corpus-level BLEU score. Figure 4 and 5 shows the BLEU score of different model by training epoch.

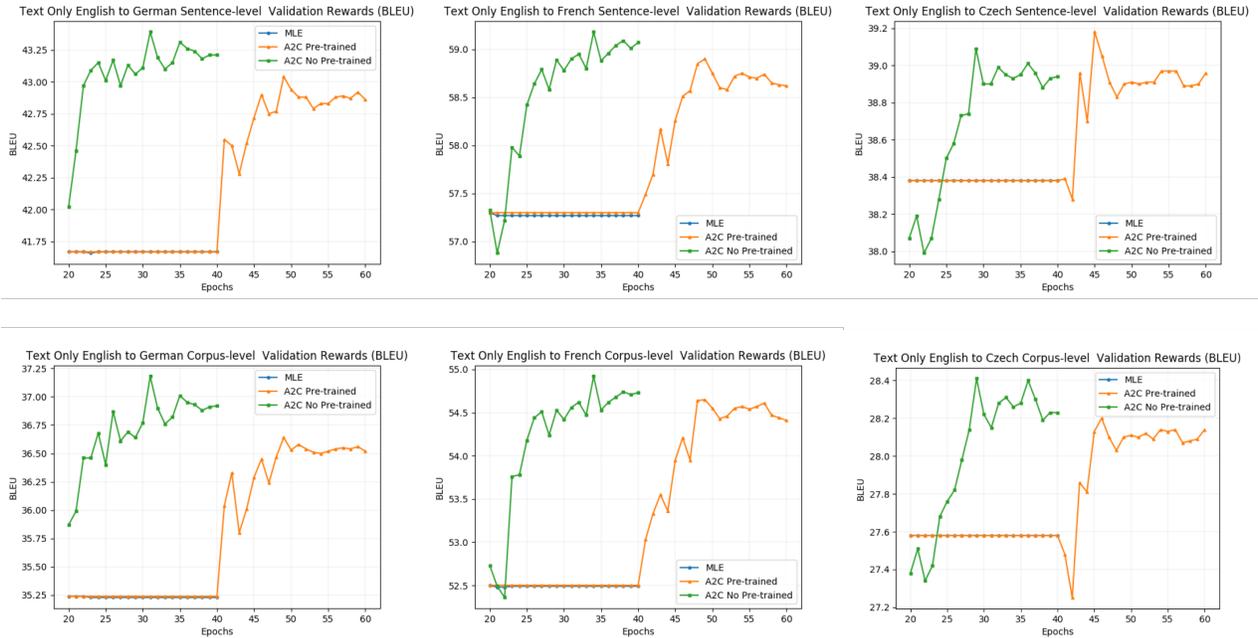

**Figure 4:** Using the architecture that has text channel only, this plot is the sentence-level (first line) and corpus-level (second line) reward curves throughout training epochs for English to German, French and Czech. Each plot contains three curves: blue curve represents MLE training from epoch 1-40; orange curve represents A2C that pre-trains actor for epoch 1-20, pretrains critic for epoch 21-40, and joint training for epoch 41-60; green curve represents A2C that only pre-trains actor for epoch 1-20, and joint training for epoch 21-40. Two more details: (1) We omit epoch 1-20 because all three curves look similar (pre-training actor uses MLE as well); (2) **Orange curve has 20 more epochs**, but we consider this as a fair comparison to green curve and blue curve, since we fixed-actor (actual performance) while pre-training critic during epoch 21-40.

### 4.4.2 Effect of pre-training critic

For text-only translation, pre-training critic is more desirable for English-German and English-French translation but not for English-Czech translation. We suspect this observation is language specific. Although English, German, French and Czech are all from Indo-European languages, English and German are from Germanic language family, while French is from Romance language family, and Czech is from Slavic language family. In addition, historically English and French borrow works and grammars from each other. Therefore, English-Czech are the most different language pair among the three (so can we see from the perplexity).

In addition, for English-French translation, pre-trained critic helps more than English-German translation. While for translation with image channel, pre-trained critic has less fluctuations during training and achieve better result, more or less, than its counterpart without pre-training, especially in English-German translation.

### 4.4.3 Effect of reinforcement A2C vs. supervised MLE

MLE plateau quickly while reinforcement A2C still grows after 20 epochs. For both text-only and text-plus-image translation, and for three language pairs, A2C overcomes the plateau problem and eventually have a higher corpus-level BLEU score.

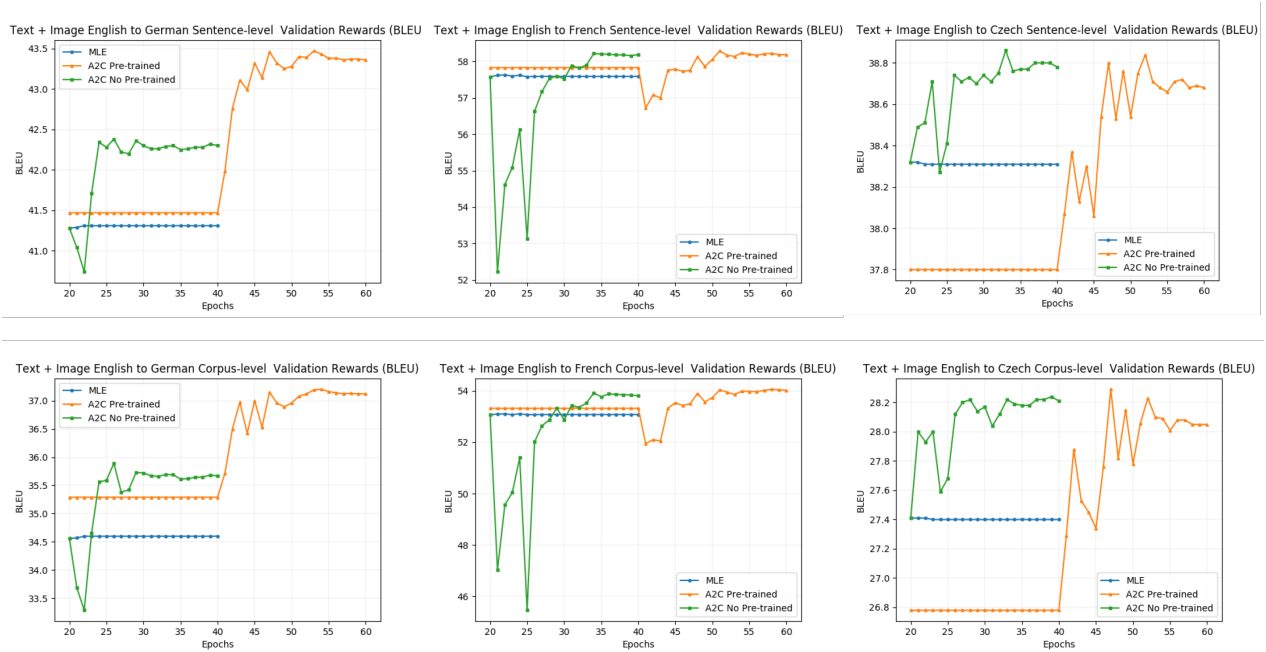

**Figure 5:** Text and Image channel architecture sentence-level (first line) and corpus-level (second line) reward curves throughout training epochs for English to German, French and Czech.

#### 4.4.4 Effect of Image channel

It looks that there is not much effect in BLEU score from the table. However, from some sample translation pair, we'll see in later sections, Image channels does make a difference. The reason why image channel doesn't help with the overall BLEU score performance could be:

- The image feature is too long (2048 dimensional), and in the MLP layer, it dilutes the output of the encoder.

- The image feature is extracted via ResNet, not trained End-to-End, containing features of objects in the image. However, minor objects are also extracted but does not appear in the image caption. Therefore, the ResNet extracted features contains noises.

- Currently we input image features into the decoder. This is simple but we could also try to input the image features into the encoder.

### 4.5 Sample translation

Table 3 shows three example sentences translated using MLE and A2C, using text-only or using image along with text. We see MLE is prone to repeating by itself, e.g. *with her hand, with her hand,* and exposure bias, e.g. *man in military* followed by *wearing a science uniform*. A2C is able to capture more poetic language, regardless of the simplest language usage pattern.

### 4.6 Model size and training speed

Table 4 and 5 shows the number of parameters and training time for each model. A2C has almost twice as much parameters as MLE, and therefore runs slower than MLE training. However, the comparably slower training speed is worthwhile for the boost in evluation metrics.

## 5 Limitation and Future Work

**Model structure improvement:** Currently, we input the extracted image features into the Decoder. We could try to input th e image features into the encoder.

**Table 3:** Sample French-English (the reversed language pair for readability) for MLE and A2C training, where we see A2C makes more consistent translation as human-captioned translation reference.

| Source | *une personne sur une motoneige en plein saut .* |
|---|---|
| Reference | *a person on a snowmobile in mid jump .* |
| MLE (text only) | *a person on an inlet on a jump .* |
| A2C (text only) | *a person on snowmobile turn .* |
| MLE (text + image) | *a person on a snowmobile at a jump .* |
| A2C (text +image) | *a person on a snowmobile in a jump .* |

| Source | *une femme en manteau rouge , avec un sac à main bleuté probablement d'apos origine asiatique , sautant en l'apos air pour une photo .* |
|---|---|
| Reference | *a lady in a red coat , holding a bluish hand bag likely of asian descent , jumping off the ground for a snapshot .* |
| MLE (text only) | *a woman in a red coat , with a purse , says have asian descent on the air to a picture .* |
| A2C (text only) | *a woman in a red coat , with a purse of her bunny purse , jumping in the air for a picture .* |
| MLE (text + image) | *a woman in a red coat , with a surgical print purse , , jumping in the air for a picture .* |
| A2C (text +image) | *a woman in a red coat , with a designer purse , presumably asian descent , jumping in the air for a picture .* |

| Source | *assise nonchalamment dans un lieu public , une fille lit en tenant le livre ouvert avec sa main , sur laquelle se trouve une bague papillon .* |
|---|---|
| Reference | *sitting casually in a public place , a girl reads holding the book open with her hand on which is a butterfly ring .* |
| MLE (text only) | *sitting in a public place , a girl reads the book with her hand , with her hand , about to a butterfly .* |
| A2C (text only) | *two sitting in a public place , a girl reads holding the book with her hand , about to a butterfly bucks .* |
| MLE (text + image) | *man in military in an audience , a girl reads while holding the open book with his hand , on which is wearing a science uniform .* |
| A2C (text +image) | *man sitting in a public place , a girl reads the open book with his hand , on wine is wearing a bowtie .* |

**Table 4:** Example of training time for text-only English to German translation

| Text-only | # parameters | sec/epoch |
|---|---|---|
| MLE (supervised) / Pre-train actor | 13,601,574 | 25.17 |
| Pre-train critic | 22,390,823 | 48.53 |
| A2C | | 65.01 |

| Text and image channel | # parameters | sec/epoch |
|---|---|---|
| MLE (supervised) / Pre-train actor | 23,043,110 | 27.72 |
| Pre-train critic | 41,273,895 | 52.14 |
| A2C | | 69.79 |

**Image features:** Currently, the image features are extracted via ResNet50. 2048 dimension of image features is too long that dilutes the text features. And the feature might contain noise which is minor objects in the image. In the future, we could try to use raw image as input and train the model end-to-end.

**Different reward signals:** Currently in A2C training, BLEU score is used as the reward signal. But there are other signals we could try, e.g. ROUGE, variants of Kappa, Error rate, overlapping precision (recall, F-1) etc. And we can use linear combination of different metrics or use functions of those metrics, e.g. log(BLEU). It would be interesting to see how different metrics could effect the A2C training.

# References

[1] D. Bahdanau, P. Brakel, K. Xu, A. Goyal, R. Lowe, J. Pineau, A. Courville, and Y. Bengio. An actor-critic algorithm for sequence prediction. *arXiv preprint arXiv:1607.07086*, 2016.

[2] S. Bengio, O. Vinyals, N. Jaitly, and N. Shazeer. Scheduled sampling for sequence prediction with recurrent neural networks. In *Advances in Neural Information Processing Systems*, pages 1171–1179, 2015.

[3] I. Calixto, D. Elliott, and S. Frank. Dcu-uva multimodal mt system report. In *Proceedings of the First Conference on Machine Translation: Volume 2, Shared Task Papers*, volume 2, pages 634–638, 2016.

[4] D. Elliott, S. Frank, K. Sima'an, and L. Specia. Multi30k: Multilingual english-german image descriptions. In *Proceedings of the 5th Workshop on Vision and Language*, pages 70–74, 2016.

[5] S. R. Guasch and M. R. Costa-Jussà. Wmt 2016 multimodal translation system description based on bidirectional recurrent neural networks with double-embeddings. In *Proceedings of the First Conference on Machine Translation: Volume 2, Shared Task Papers*, volume 2, pages 655–659, 2016.

[6] D. He, Y. Xia, T. Qin, L. Wang, N. Yu, T. Liu, and W.-Y. Ma. Dual learning for machine translation. In *Advances in Neural Information Processing Systems*, pages 820–828, 2016.

[7] K. He, X. Zhang, S. Ren, and J. Sun. Deep residual learning for image recognition. In *Proceedings of the IEEE conference on computer vision and pattern recognition*, pages 770–778, 2016.

[8] S. Jean, O. Firat, K. Cho, R. Memisevic, and Y. Bengio. Montreal neural machine translation systems for wmt'15. In *Proceedings of the Tenth Workshop on Statistical Machine Translation*, pages 134–140, 2015.

[9] K. Papineni, S. Roukos, T. Ward, and W.-J. Zhu. Bleu: a method for automatic evaluation of machine translation. In *Proceedings of the 40th annual meeting on association for computational linguistics*, pages 311–318. Association for Computational Linguistics, 2002.

[10] M. Ranzato, S. Chopra, M. Auli, and W. Zaremba. Sequence level training with recurrent neural networks. *arXiv preprint arXiv:1511.06732*, 2015.

[11] L. Specia, S. Frank, K. Sima'an, and D. Elliott. A shared task on multimodal machine translation and crosslingual image description. In *Proceedings of the First Conference on Machine Translation: Volume 2, Shared Task Papers*, volume 2, pages 543–553, 2016.

[12] O. Vinyals, A. Toshev, S. Bengio, and D. Erhan. Show and tell: A neural image caption generator. In *Computer Vision and Pattern Recognition (CVPR), 2015 IEEE Conference on*, pages 3156–3164. IEEE, 2015.

[13] K. Xu, J. Ba, R. Kiros, K. Cho, A. Courville, R. Salakhudinov, R. Zemel, and Y. Bengio. Show, attend and tell: Neural image caption generation with visual attention. In *International Conference on Machine Learning*, pages 2048–2057, 2015.